\documentclass[runningheads]{llncs}
\usepackage[T1]{fontenc}
\usepackage{graphicx}
\usepackage{subcaption}
\usepackage{booktabs}
\usepackage{multirow}
\usepackage{array}
\usepackage{xcolor}
\usepackage{makecell}
\usepackage{pifont}
\usepackage{amsmath}
\usepackage{hyperref}
\usepackage{orcidlink}
\usepackage{amssymb} 

\begin{document}
\title{TRACE: Temporal Radiology with Anatomical Change Explanation for Grounded X-ray Report Generation}
\titlerunning{TRACE}
\author{OFM Riaz Rahman Aranya\orcidlink{0000-0002-8195-2710} \and
Kevin Desai\orcidlink{0000-0002-2964-8981}}
\authorrunning{O. R. R. Aranya \and K. Desai}
\institute{The University of Texas at San Antonio, TX 78249, USA \\
\email{\{ofmriazrahman.aranya, kevin.desai\}@utsa.edu}}
\maketitle
\begin{abstract}
Temporal comparison of chest X-rays is fundamental to clinical radiology, enabling detection of disease progression, treatment response, and new findings. While vision-language models have advanced single-image report generation and visual grounding, no existing method combines these capabilities for temporal change detection. We introduce Temporal Radiology with Anatomical Change Explanation (TRACE), the first model that jointly performs temporal comparison, change classification, and spatial localization. Given a prior and current chest X-ray, TRACE generates natural language descriptions of interval changes (worsened, improved, stable) while grounding each finding with bounding box coordinates. TRACE demonstrates effective spatial localization with over 90\% grounding accuracy, establishing a foundation for this challenging new task. Our ablation study uncovers an emergent capability: change detection arises only when temporal comparison and spatial grounding are jointly learned, as neither alone enables meaningful change detection. This finding suggests that grounding provides a spatial attention mechanism essential for temporal reasoning.

Code is available at 
\href{https://github.com/UTSA-VIRLab/TRACE}{https://github.com/UTSA-VIRLab/TRACE}

\end{abstract}

\section{Introduction}
\label{sec:introduction}
Radiologists routinely compare current chest X-rays with prior studies to detect interval changes, determining whether findings have worsened, improved, or remained stable. These temporal assessments are critical for clinical decision-making: a worsening pneumothorax may require immediate intervention, while an improving pleural effusion suggests effective treatment. Studies have demonstrated that longitudinal chest X-ray analysis provides important insights into disease progression, guides treatment decisions, and improves prognostic accuracy~\cite{duong2023longitudinal,li2020siamese}. In the MIMIC-CXR~\cite{johnson2019mimic} dataset alone, more than 67\% of patients undergo multiple examinations, yet existing automated methods largely ignore this temporal information.

\begin{figure*}[t]
    \centering
    \includegraphics[width=\textwidth]{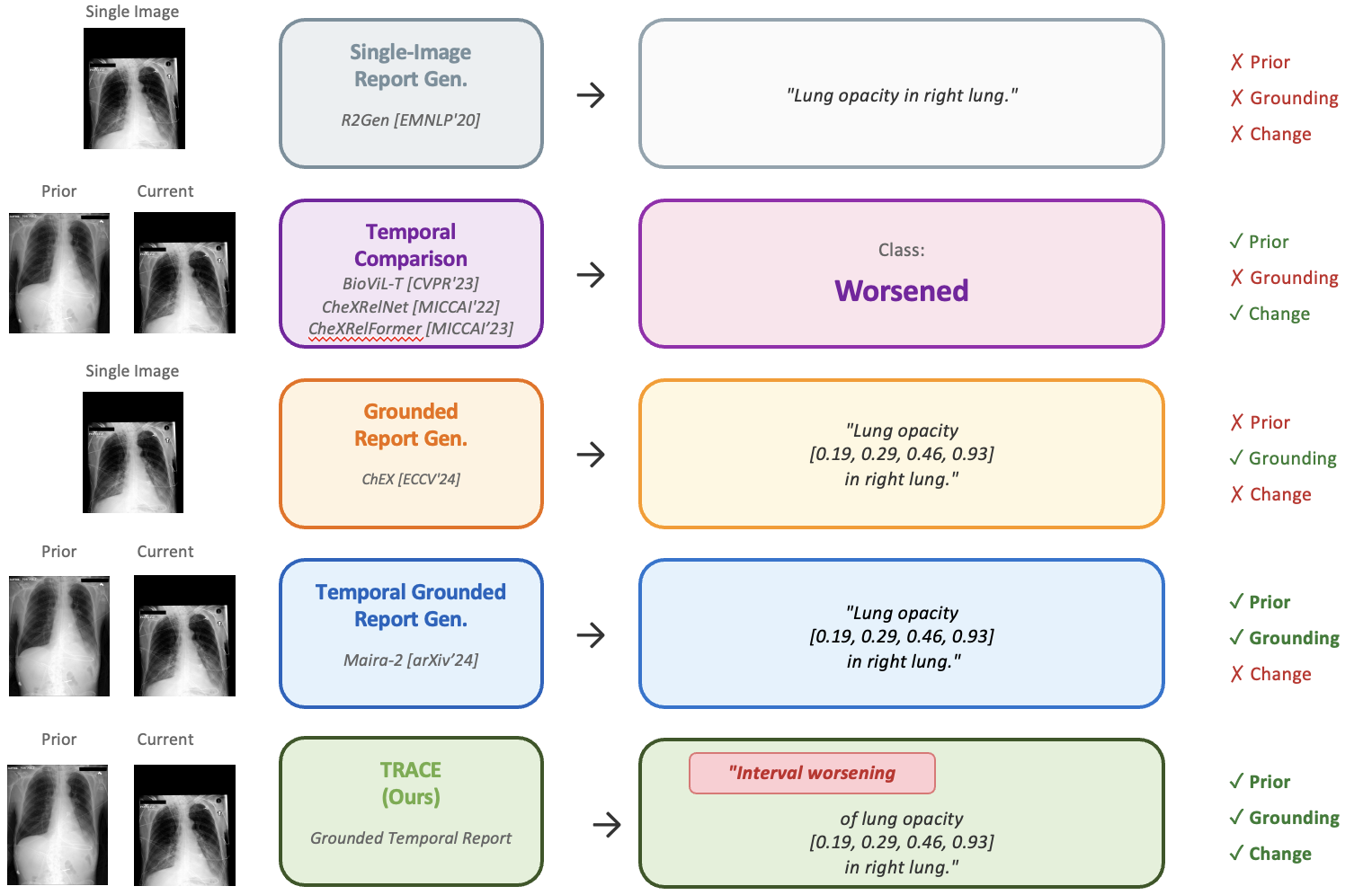}
    \caption{Comparison of chest X-ray report generation approaches. 
    \textbf{Single-Image Report Generation} (R2Gen~\cite{chen2020generating}) 
    generates reports without temporal context or spatial grounding. 
    \textbf{Temporal Comparison} (BioViL-T~\cite{bannur2023learning}, 
    CheXRelNet~\cite{karwande2022chexrelnet}, CheXRelFormer~\cite{mbakwe2023chexrelformer}) 
    detects disease progression but outputs only classification labels. 
    \textbf{Grounded Report Generation} (ChEX~\cite{muller2024chex}) produces 
    spatially grounded reports but lacks temporal reasoning. 
    \textbf{Temporal Grounded Report Generation} (MAIRA-2~\cite{bannur2024maira2}) 
    uses prior images with grounding but without explicit change classification. 
    \textbf{TRACE (Ours)} is the first to combine explicit temporal change 
    classification with spatial grounding, describing what changed, where, and how.}
    \label{fig:intro_comparison}
\end{figure*}

Despite the clinical importance of temporal comparison, automated radiology report generation methods have predominantly focused on single-image analysis~\cite{jing2018automatic,chen2020generating,miura2021improving,chen2021cross}. These approaches, while achieving impressive results on individual images, cannot capture disease progression or generate the comparison statements that are central to clinical radiology reports~\cite{hyland2023maira}. Recent surveys highlight this gap, noting that without access to prior studies, models often hallucinate spurious temporal references or fail to capture meaningful changes~\cite{messina2022survey}.

Recent advances in vision-language models have enabled substantial progress in medical image analysis. Encoder-decoder architectures with attention mechanisms~\cite{chen2020generating}, memory-driven transformers~\cite{chen2021cross}, and knowledge graph integration~\cite{zhang2020radiology,li2019knowledge} have improved report quality. Large language models have further advanced the field, enabling conversational radiology assistance~\cite{pellegrini2023radialog} and more natural report generation~\cite{liu2024visual,li2024llavamed}. Separately, visual grounding methods have emerged to localize findings with bounding boxes, including MS-CXR~\cite{boecking2022making}, medical phrase grounding approaches~\cite{chen2023medrpg}, and grounded report generation systems~\cite{bannur2024maira2,muller2024chex}. These grounding methods improve interpretability by showing where findings are located, but operate on single images without temporal context. Conversely, methods that do address temporal comparison lack the ability 
to spatially localize where changes occur.

Some recent work has begun to address temporal aspects of chest X-ray analysis. BioViL-T~\cite{bannur2023learning} introduced temporal pre-training to learn representations that capture changes across studies. CheXRelNet~\cite{karwande2022chexrelnet} proposed tracking longitudinal pathology relationships between chest X-ray pairs. MAIRA-2~\cite{bannur2024maira2} uses prior images and reports as context for current report generation, improving consistency. Methods incorporating longitudinal data have shown improved clinical accuracy~\cite{peng2023longitudinal,liu2025mlrg}. However, none of these approaches combines explicit temporal change classification (worsened/improved/stable) with spatial grounding to localize where changes occur.

We introduce TRACE (Temporal Radiology with Anatomical Change Explanation), the first model for grounded temporal change detection in chest X-rays. 
As illustrated in Figure~\ref{fig:intro_comparison}, existing methods address only subsets of this problem: single-image report generation 
lacks temporal context, temporal classification lacks spatial grounding, 
and grounded generation lacks temporal reasoning. TRACE unifies all three 
capabilities. Given a prior and current study, TRACE generates a natural language report describing interval changes while localizing each finding with bounding box coordinates. For example, given two chest X-rays, TRACE produces outputs such as: \texttt{``Interval worsening of pneumothorax <box>0.19,0.11,0.52,0.63</box> in right lung.''} This unified output provides clinically actionable information: the finding (pneumothorax), the change (worsening), and the location (bounding box).

Our ablation study reveals a surprising finding: temporal change detection is an emergent capability that requires both temporal comparison and spatial grounding. Remarkably, a model with access to both prior and current images but without grounding supervision fails at change detection, predicting ``stable'' for all samples. We hypothesize that grounding provides a spatial attention mechanism that enables the model to focus on specific anatomical regions when comparing images temporally.

Our contributions are:
\begin{itemize}
\item We introduce grounded temporal change detection, a new task combining temporal comparison, change classification, and spatial localization in a unified framework.
\item We discover an emergent capability: change detection only arises when both temporal comparison and spatial grounding are combined. Without grounding supervision, models fail to detect changes even with access to both images.
\item We construct a large-scale dataset with grounded temporal annotations from MIMIC-CXR and Chest ImaGenome (79,202 training, 22,553 test samples) to enable systematic evaluation of temporal reasoning with spatial grounding.
\end{itemize}

\section{Related Work}
\label{sec:related_work}

\textbf{Radiology Report Generation.} Automated radiology report generation has evolved significantly from early template-based approaches to sophisticated deep learning methods. The encoder-decoder paradigm dominates the field, where visual encoders extract features from medical images and language decoders generate textual reports~\cite{jing2018automatic,chen2020generating}. Jing et al.~\cite{jing2018automatic} employed CNN encoders with LSTM decoders and co-attention mechanisms to produce coherent reports. The introduction of Transformer-based architectures marked a significant advance, with R2Gen~\cite{chen2020generating} and R2GenCMN~\cite{chen2021cross} improving the capture of long-range dependencies and reducing repetitive text generation. Knowledge graph integration~\cite{zhang2020radiology,li2019knowledge} injects medical domain knowledge to generate more clinically relevant descriptions. Miura et al.~\cite{miura2021improving} applied reinforcement learning with clinical metrics as rewards to improve factual correctness. METransformer~\cite{wang2023metransformer} captures fine-grained visual patterns through learnable expert tokens. Despite these advances, most methods process single images and cannot 
generate the temporal comparison statements that are fundamental to 
clinical radiology practice (Figure~\ref{fig:intro_comparison}, Row 1).

\textbf{Temporal Analysis in Chest Radiography.} Longitudinal analysis of chest X-rays is essential for monitoring disease progression and treatment response~\cite{duong2023longitudinal,li2020siamese}. Li et al.~\cite{li2020siamese} employed Siamese neural networks to evaluate disease severity changes between longitudinal patient visits using metric learning. CheXRelNet~\cite{karwande2022chexrelnet} tracks pathology change relations between chest X-ray pairs using anatomy-aware graph representations on the Chest ImaGenome dataset. For report generation, BioViL-T~\cite{bannur2023learning} introduced temporal pre-training objectives that learn to predict whether findings have progressed, improved, or remained stable. Peng et al.~\cite{peng2023longitudinal} utilized longitudinal data for report pre-filling, while Liu et al.~\cite{liu2025mlrg} proposed multi-view longitudinal contrastive learning to improve clinical accuracy. MAIRA-2~\cite{bannur2024maira2} uses prior images and reports as context 
inputs to reduce hallucinations and improve consistency with previous findings,
and additionally supports spatial grounding (Figure~\ref{fig:intro_comparison}, Row 4).
However, these approaches either perform classification without localization 
or generate reports without explicit change classification, leaving a gap for unified 
grounded temporal change detection.

\textbf{Visual Grounding in Medical Imaging.} Visual grounding aims to localize findings mentioned in text within corresponding image regions. In natural images, this task has been extensively studied, but medical imaging presents unique challenges due to subtle pathology appearances and specialized anatomical terminology. MS-CXR~\cite{boecking2022making} introduced a benchmark dataset pairing chest X-ray images with phrase-level bounding box annotations, enabling the development of medical phrase grounding methods. MedRPG~\cite{chen2023medrpg} proposed an end-to-end transformer-based approach for medical phrase grounding with region-phrase contrastive alignment. ChEX~\cite{muller2024chex} introduced interactive localization supporting both textual queries and bounding box inputs for chest X-ray interpretation. MAIRA-2~\cite{bannur2024maira2} extended grounding to report generation, 
producing reports where individual findings are associated with spatial 
annotations. While MAIRA-2 incorporates prior images to reduce hallucinations, 
it does not explicitly classify temporal changes (worsened/improved/stable). 
ChEX and other single-image grounding methods operate without temporal 
context (Figure~\ref{fig:intro_comparison}, Row 3).

\textbf{Vision-Language Models in Medicine.} Large vision-language models (VLMs) have transformed medical image analysis by combining visual understanding with natural language capabilities. LLaVA~\cite{liu2024visual} introduced visual instruction tuning, which has been adapted to the medical domain through LLaVA-Med~\cite{li2024llavamed}, trained on biomedical image-text pairs from PubMed. CheXagent~\cite{chen2024chexagent} developed a foundation model specifically for chest X-ray interpretation supporting multiple tasks including report generation and visual question answering. RaDialog~\cite{pellegrini2023radialog} extended medical VLMs to support conversational radiology assistance with interactive dialogue capabilities. MAIRA-1~\cite{hyland2023maira} demonstrated that specialized radiology models can outperform larger general-purpose models. Despite rapid progress, existing medical VLMs primarily operate on single images and lack explicit mechanisms for temporal change detection with spatial grounding. Our work addresses this gap by unifying temporal comparison, change 
classification, and spatial localization into a single framework 
(Figure~\ref{fig:intro_comparison}, Row 5).

\section{Method}
\label{sec:method}

\subsection{Dataset Construction}
\label{sec:dataset}

We construct our temporal grounding dataset from two publicly available sources: MIMIC-CXR-JPG~\cite{johnson2019mimic} for chest X-ray images and Chest ImaGenome~\cite{wu2021chest} for anatomical annotations and temporal comparison labels.

\textbf{Data Sources.} MIMIC-CXR-JPG contains 377,110 chest radiographs from 65,379 patients at Beth Israel Deaconess Medical Center. Importantly, many patients have multiple studies over time, enabling temporal analysis. Chest ImaGenome provides automatically extracted scene graphs for 242,072 MIMIC-CXR images, including bounding boxes for 29 anatomical regions and temporal comparison annotations indicating whether findings have improved, worsened, or remained stable compared to prior studies.

\textbf{Temporal Pair Creation.} We create training samples by pairing consecutive studies from the same patient. For each patient with multiple examinations, we: (1) group all studies by patient ID, (2) sort studies chronologically using the StudyOrder metadata, and (3) pair each study with its immediate predecessor to form (prior, current) image pairs. We only include pairs where the current study contains at least one temporal comparison annotation.

\textbf{Annotation Format.} For each image pair, we extract temporal annotations from the Chest ImaGenome scene graphs and format them as grounded sentences:
\begin{quote}
\small
\texttt{``Interval worsening of pneumothorax <box>0.196,0.107,0.522,0.634</box> in right lung.''}
\end{quote}

\textbf{Dataset Statistics.} Table~\ref{tab:dataset} summarizes our dataset. We use the official patient-disjoint splits from Chest ImaGenome to maintain strict separation between training and evaluation cohorts.

\begin{table}[h]
\centering
\caption{Dataset statistics for TRACE.}
\label{tab:dataset}
\begin{tabular}{lccc}
\toprule
& \textbf{Train} & \textbf{Val} & \textbf{Test} \\
\midrule
Samples & 79,202 & 11,464 & 22,553 \\
\midrule
\textit{Change Distribution} & & & \\
\quad Worsened & 27.7\% & 28.4\% & 34.5\% \\
\quad Improved & 20.1\% & 20.1\% & 21.7\% \\
\quad Stable & 52.2\% & 51.5\% & 43.8\% \\
\bottomrule
\end{tabular}
\end{table}

\subsection{Model Architecture}
\label{sec:architecture}

\begin{figure*}[h]
\vspace{-3mm}
\centering
\includegraphics[width=\textwidth]{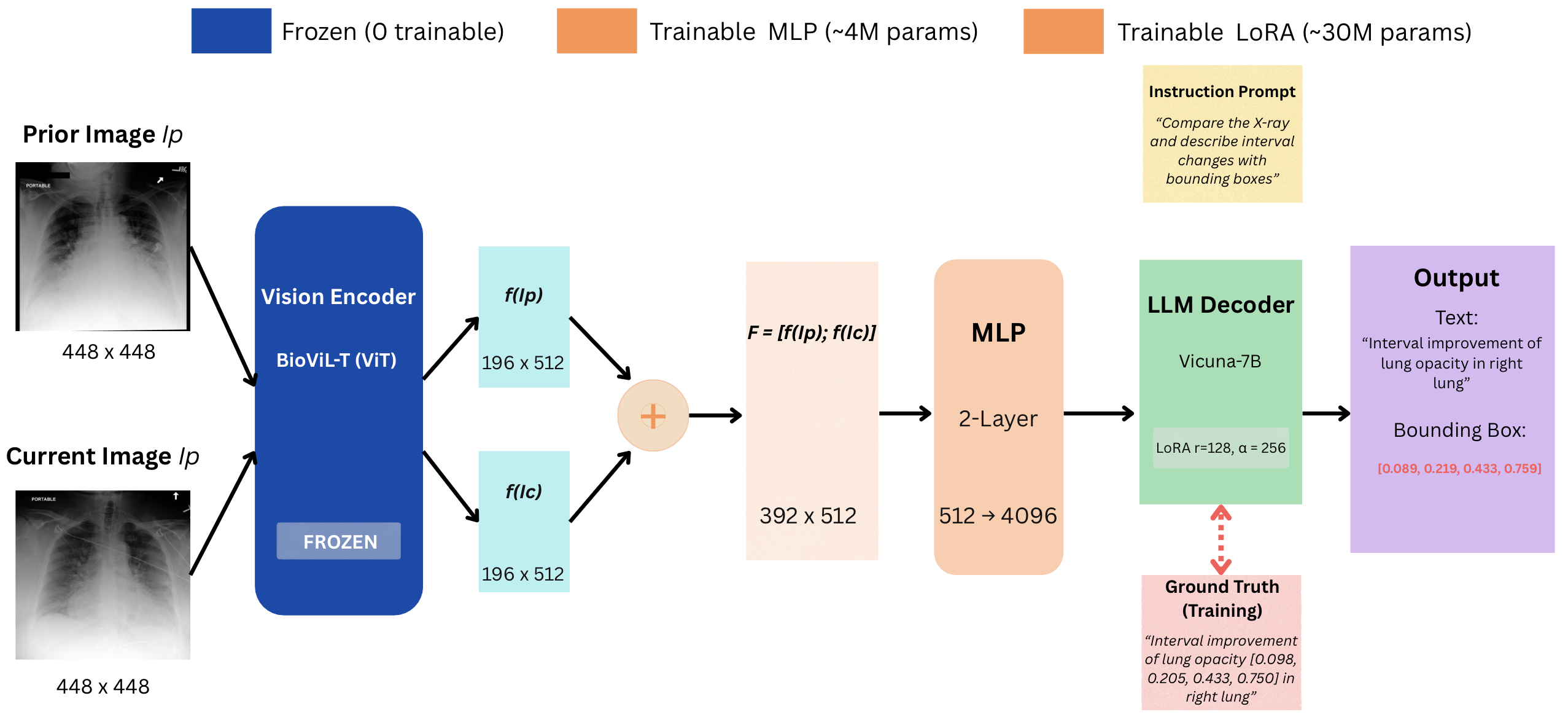}
\caption{Overview of the TRACE architecture. Prior and current chest X-rays are encoded separately by a frozen BioViL-T encoder with shared weights. The resulting feature sequences (196 tokens each) are concatenated to form 392 visual tokens, projected via a trainable MLP, and decoded by a large language model with LoRA fine-tuning to generate grounded temporal reports.}
\label{fig:architecture}
\vspace{-3mm}
\end{figure*}

TRACE builds upon the LLaVA~\cite{liu2024visual} framework, adapting it for temporal comparison of chest X-rays. Figure~\ref{fig:architecture} illustrates our architecture.

\textbf{Vision Encoder.} We employ BioViL-T~\cite{bannur2023learning}, a vision transformer pre-trained on chest X-rays with temporal awareness. Given an input image resized to 448×448 pixels, BioViL-T produces a sequence of 196 visual tokens (14 × 14 spatial grid with total stride 32), each with dimension 512. The encoder remains frozen during training to preserve learned radiological representations.

\textbf{Temporal Feature Fusion.} We employ feature concatenation followed by implicit cross-attention in the language model rather than explicit temporal attention modules or difference images. Given a prior image $I_p$ and current image $I_c$, we extract features independently using shared encoder weights and concatenate along the sequence dimension:
\begin{equation}
    F = f(I_p) \oplus f(I_c)
\end{equation}
This produces 392 visual tokens that are jointly processed by the language model. The self-attention mechanism in the decoder enables implicit comparison between corresponding spatial regions across both images, allowing the model to identify temporal changes without lossy operations like image subtraction. This design preserves complete information from both studies while leveraging the language model's capacity for relational reasoning.

\textbf{MLP Projector.} A two-layer MLP with GELU activation projects the 512-dimensional visual features to the 4096-dimensional embedding space of the language model, introducing approximately 4M trainable parameters.

\textbf{Language Model Decoder.} We use Vicuna-7B~\cite{chiang2023vicuna} as our primary decoder, which autoregressively generates grounded temporal reports. The output includes natural language change descriptions with embedded bounding box coordinates in normalized format. We apply Low-Rank Adaptation (LoRA)~\cite{hu2022lora} to the query and value projection matrices in all attention layers with rank $r=128$ and scaling factor $\alpha=256$, introducing approximately 30M trainable parameters while keeping the base model frozen. We also evaluate Mistral-7B~\cite{jiang2023mistral} as an alternative backbone in Section~\ref{sec:experiments}. Table~\ref{tab:trainable} summarizes the trainable components.

\begin{table}[h]
\vspace{-6mm}
\centering
\caption{Trainable components in TRACE.}
\label{tab:trainable}
\begin{tabular}{lcc}
\toprule
\textbf{Component} & \textbf{Status} & \textbf{Parameters} \\
\midrule
BioViL-T Encoder & Frozen & 0 \\
MLP Projector & Trained & $\sim$4M \\
LLM (LoRA) & Fine-tuned & $\sim$30M \\
\midrule
\textbf{Total Trainable} & & $\sim$34M \\
\bottomrule
\end{tabular}
\vspace{-6mm}
\end{table}

\subsection{Training}
\label{sec:training}

We adopt a two-stage training procedure following the LLaVA framework~\cite{liu2024visual}. In the first stage, we train only the MLP projector to align the visual features from BioViL-T with the language model's embedding space, keeping both the vision encoder and language model frozen. This alignment stage uses a subset of our dataset for one epoch. In the second stage, we perform end-to-end instruction tuning where the MLP projector and language model are jointly optimized while the vision encoder remains frozen. The model learns to generate grounded temporal reports through standard autoregressive language modeling, where the training objective minimizes the negative log-likelihood of the target tokens conditioned on the visual features and instruction prompt. To enable efficient adaptation with limited computational resources, we employ Low-Rank Adaptation (LoRA)~\cite{hu2022lora} for the language model, applying adapters to the query and value projection matrices in all attention layers with rank $r=128$ and scaling factor $\alpha=256$. This configuration introduces approximately 34M trainable parameters while preserving the pre-trained knowledge of the base model.

\section{Experiments}
\label{sec:experiments}

\textbf{Experimental Setup.} We evaluate TRACE on our temporal grounding benchmark comprising 22,553 test samples, each consisting of a prior-current image pair with grounded temporal annotations. The vision encoder is initialized from BioViL-T~\cite{bannur2023learning}, and the language model from Vicuna-7B~\cite{chiang2023vicuna}. Since no prior work combines explicit temporal change classification with spatial grounding, we validate TRACE through comprehensive ablation studies that isolate the contribution of each component.

\textbf{Evaluation Metrics.} We evaluate performance across four dimensions: (1) change detection accuracy measuring three-way classification (worsened, improved, stable), (2) grounding accuracy using Intersection over Union (IoU) between predicted and ground truth bounding boxes, (3) natural language generation metrics including BLEU-4, METEOR, and ROUGE-L for text quality, and (4) clinical efficacy metrics including RadGraph F1~\cite{jain2021radgraph} and CheXbert F1~\cite{smit2020chexbert} for clinical correctness.

\textbf{Main Results.} Table~\ref{tab:main_results} presents comprehensive 
results across all evaluation metrics. TRACE achieves 48.0\% change detection 
accuracy on the challenging three-way classification task. As our ablation 
study demonstrates (Table~\ref{tab:ablation}), this capability emerges only 
when both temporal input and grounding supervision are present; removing 
either causes complete failure, with models defaulting to predicting ``stable'' 
for all samples. For grounding, TRACE achieves 90.2\% accuracy at IoU$>$0.5, 
demonstrating precise spatial localization of temporal changes. Report 
generation metrics show strong performance on both NLG and clinical efficacy 
measures, with RadGraph and CheXbert F1 scores confirming clinically accurate outputs.

\begin{table}[h]
\vspace{-8mm}
\centering
\caption{TRACE performance on the test set (22,553 samples).}
\label{tab:main_results}
\begin{tabular}{llc}
\toprule
\textbf{Category} & \textbf{Metric} & \textbf{Value} \\
\midrule
\multirow{3}{*}{NLG Metrics} 
& BLEU-4 & 0.260 \\
& METEOR & 0.438 \\
& ROUGE-L & 0.494 \\
\midrule
\multirow{2}{*}{Clinical Efficacy}
& RadGraph F1 & 0.406 \\
& CheXbert F1 & 0.432 \\
\midrule
\multirow{2}{*}{Grounding}
& Mean IoU & 0.772 \\
& IoU $>$ 0.5 & 90.2\% \\
\midrule
Change Detection & Accuracy & 48.0\% \\
\bottomrule
\end{tabular}
\vspace{-4mm}
\end{table}

\textbf{Comparison with Prior Methods.} Table~\ref{tab:comparison} compares TRACE with existing temporal classification methods on Chest ImaGenome. CheXRelNet~\cite{karwande2022chexrelnet} uses anatomy-aware graph representations to predict change relations, achieving 46.8\% accuracy. CheXRelFormer~\cite{mbakwe2023chexrelformer} improves upon this with hierarchical vision transformers, reaching 49.3\% accuracy. However, both methods are limited to classification only. They cannot localize where changes occur and cannot generate natural language descriptions.
TRACE achieves comparable temporal accuracy (48.0\%) while providing two additional capabilities: spatial grounding with 90.2\% accuracy at IoU$>$0.5, and full natural language report generation. This unified output describes what changed, where, and how, providing clinically actionable information that pure classification methods cannot offer.

\begin{table}[t]
\centering
\caption{Comparison with prior temporal classification methods on Chest ImaGenome.}
\label{tab:comparison}
\begin{tabular}{lccc}
\toprule
\textbf{Method} & \textbf{Temporal Acc.} & \textbf{Grounding} & \textbf{Report Gen} \\
\midrule
CheXRelNet~\cite{karwande2022chexrelnet} & 46.8\% & \texttimes & \texttimes \\
CheXRelFormer~\cite{mbakwe2023chexrelformer} & 49.3\% & \texttimes & \texttimes \\
\textbf{TRACE (Ours)} & 48.0\% & \checkmark (90.2\%) & \checkmark \\
\bottomrule
\end{tabular}
\vspace{-4mm}
\end{table}

\textbf{Per-Class Analysis.} Table~\ref{tab:perclass} presents per-class change detection performance. Stable findings achieve the highest recall (67.4\%) as the most frequent class, while improvement detection proves most challenging with 26.3\% recall. This difficulty likely stems from resolving pathologies leaving subtle residual findings that are visually similar to stable conditions. Worsening detection achieves moderate performance with 37.5\% recall, suggesting that progressive deterioration produces more distinctive visual signatures than gradual improvement.

\begin{table}[h]
\vspace{-8mm}
\centering
\caption{Per-class change detection performance.}
\label{tab:perclass}
\begin{tabular}{lcccc}
\toprule
\textbf{Class} & \textbf{Precision} & \textbf{Recall} & \textbf{F1} & \textbf{Support} \\
\midrule
Worsening & 0.485 & 0.375 & 0.423 & 7,787 \\
Improvement & 0.382 & 0.263 & 0.311 & 4,888 \\
Stable & 0.505 & 0.674 & 0.578 & 9,878 \\
\midrule
Macro Average & 0.457 & 0.437 & 0.437 & 22,553 \\
\bottomrule
\end{tabular}
\vspace{-4mm}
\end{table}

\begin{table}[b]
\vspace{-8mm}
\centering
\caption{Change detection accuracy by anatomical region.$^*$}
\label{tab:anatomy}
\begin{tabular}{lcc}
\toprule
\textbf{Anatomy} & \textbf{Accuracy} & \textbf{Support} \\
\midrule
Mediastinum & 75.5\% & 339 \\
Cardiac Silhouette & 68.8\% & 1,868 \\
Right Hilar & 61.7\% & 60 \\
Right Lung & 45.3\% & 17,343 \\
Left Lung & 44.7\% & 2,549 \\
\bottomrule
\multicolumn{3}{l}{\small $^*$ Top 5 regions shown; 394 samples from other}\\
\multicolumn{3}{l}{\small \phantom{$^*$} anatomical regions omitted for brevity.}
\end{tabular}
\end{table}

\textbf{Per-Anatomy Analysis.} We analyze change detection accuracy across anatomical regions to understand model behavior on different finding types (Table~\ref{tab:anatomy}). TRACE achieves substantially higher accuracy on cardiac silhouette (68.8\%) and mediastinal changes (75.5\%) compared to lung parenchymal changes (45.3\% right lung, 44.7\% left lung). This disparity reflects the visual characteristics of different pathologies: cardiac changes involve distinct boundary shifts that are relatively easy to detect, while pulmonary changes such as opacity evolution and effusion resolution are often more diffuse and subtle. This finding suggests that future work should prioritize improving detection of subtle parenchymal changes.

\textbf{Ablation Study.} We conduct ablation experiments to understand the contribution of temporal input and grounding supervision (Table~\ref{tab:ablation}). We evaluate three variants: (1) the full TRACE model with both temporal inputs and grounding supervision, (2) a single-image variant that receives only the current image without the prior study, and (3) a no-grounding variant that receives both images but is trained without bounding box supervision. For the no-grounding variant, we remove all bounding box coordinate tokens from the training targets, training the model to generate only natural language descriptions without spatial annotations.

Removing the prior image causes a substantial drop in BLEU-4 from 0.260 to 0.051, as the model can no longer generate temporal comparison language. Change detection accuracy falls to 43.8\%, matching the majority class baseline since the model defaults to predicting ``stable'' for all samples. Interestingly, grounding accuracy remains relatively high (83.7\%), indicating that single-image localization is learnable even without temporal context.

The no-grounding variant reveals a more surprising result. Despite having access to both prior and current images, this model also fails at change detection, achieving the same 43.8\% accuracy by predicting ``stable'' for all samples. This suggests that grounding supervision provides a critical inductive bias for temporal comparison. We hypothesize that without explicit spatial localization, the model cannot effectively compare corresponding anatomical regions across time points. The grounding objective forces the model to attend to specific anatomical structures, enabling meaningful temporal reasoning.

\begin{table}[h]
\centering
\caption{Ablation study on temporal input and grounding supervision.}
\label{tab:ablation}
\begin{tabular}{lcccc}
\toprule
\textbf{Model} & \textbf{Prior Image} & \textbf{Grounding} & \textbf{BLEU-4} & \textbf{Change Accuracy} \\
\midrule
TRACE (Full) & \checkmark & \checkmark & 0.260 & 48.0\% \\
Single-Image & \texttimes & \checkmark & 0.051 & 43.8\%$^\dagger$ \\
No-Grounding & \checkmark & \texttimes & 0.240 & 43.8\%$^\dagger$ \\
\bottomrule
\multicolumn{5}{l}{\small $^\dagger$ Predicts ``stable'' for all samples (0\% recall on worsening/improvement)}
\end{tabular}
\vspace{-4mm}
\end{table}

\begin{table}[b]
\vspace{-8mm}
\centering
\caption{Language model comparison showing sensitivity-specificity trade-off.}
\label{tab:llm}
\begin{tabular}{lccccc}
\toprule
\textbf{LLM Backbone} & \textbf{Accuracy} & \makecell{\textbf{Worsening}\\\textbf{Recall}} & \makecell{\textbf{Improvement}\\\textbf{Recall}} & \makecell{\textbf{Stable}\\\textbf{Recall}} & \textbf{IoU$>$0.5} \\
\midrule
Vicuna-7B & \textbf{48.0\%} & 37.5\% & 26.3\% & \textbf{67.4\%} & \textbf{90.2\%} \\
Mistral-7B & 45.0\% & \textbf{73.6\%} & \textbf{57.9\%} & 15.7\% & 54.5\% \\
\bottomrule
\end{tabular}
\end{table}

\textbf{Language Model Comparison.} To understand the impact of language model choice, we compare Vicuna-7B and Mistral-7B~\cite{jiang2023mistral} backbones while keeping the vision encoder fixed (Table~\ref{tab:llm}). The comparison reveals distinct clinical behaviors between language models. Vicuna-7B achieves higher overall accuracy (48.0\%) through conservative prediction, favoring ``stable'' and resulting in high specificity but lower sensitivity (37.5\% worsening recall, 26.3\% improvement recall). In contrast, Mistral-7B exhibits aggressive change detection with substantially higher sensitivity (73.6\% worsening recall, 57.9\% improvement recall) but reduced specificity (15.7\% stable recall).

This sensitivity-specificity trade-off has important clinical implications. Vicuna-7B may be preferred in screening applications where false alarms are costly, while Mistral-7B may be more suitable for high-risk monitoring where missing actual changes is unacceptable. The choice of language model backbone should therefore be guided by the specific clinical requirements of the deployment scenario.

\begin{figure*}[h]
    \vspace{-3mm}
    \centering
    \includegraphics[width=\textwidth]{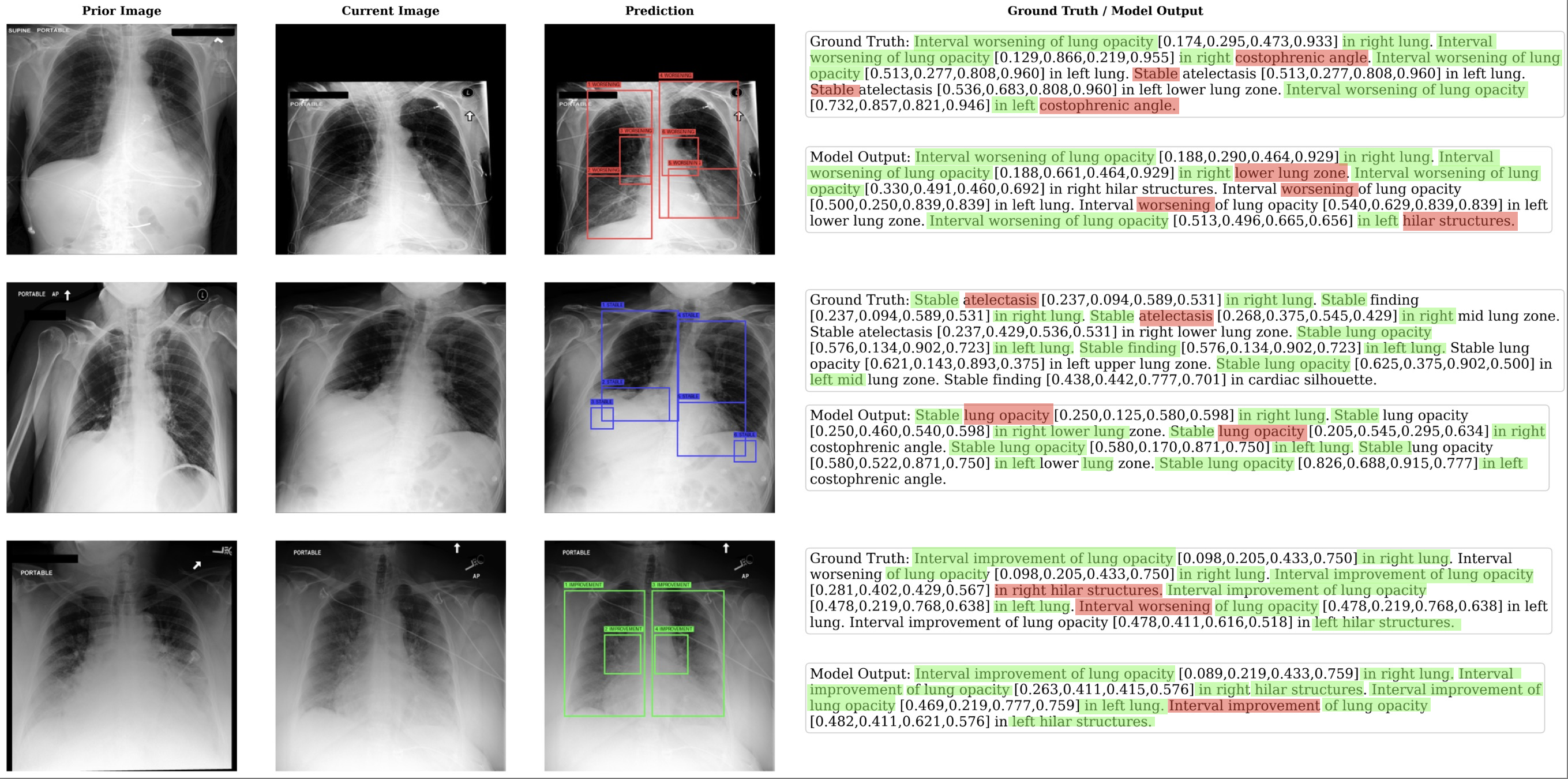}
    \caption{Qualitative results of TRACE on three representative cases: worsening, stable, and improvement. Each row shows the prior image, current image, model prediction with bounding box overlay, and the corresponding ground truth and model output text. The model successfully identifies temporal changes and localizes the affected anatomical regions with accurate bounding box coordinates.}
    \label{fig:qualitative}
    \vspace{-3mm}
\end{figure*}

\textbf{Qualitative Results.} Figure~\ref{fig:qualitative} presents representative examples of TRACE predictions across three change categories. For worsening cases, TRACE correctly identifies progressive deterioration such as increasing lung opacity, with accurate bounding box localization highlighting the affected region. For stable cases, the model appropriately identifies unchanged findings and maintains consistent spatial localization. For improvement cases, TRACE detects resolving pathologies including decreasing pleural effusion. These examples demonstrate that TRACE successfully integrates temporal reasoning with spatial grounding to produce clinically interpretable outputs.

\section{Conclusion}

We presented TRACE, the first model for grounded temporal change detection 
in chest X-rays. Given a prior and current study, TRACE generates natural 
language descriptions of interval changes while localizing each finding 
with bounding boxes. Our experiments demonstrate strong spatial localization 
with over 90\% grounding accuracy and clinically accurate report generation.
Our ablation study reveals that change detection is an emergent capability 
requiring both temporal input and grounding supervision. Models with access 
to both images but without grounding fail entirely, suggesting that spatial 
localization provides a critical inductive bias for temporal reasoning. 
Per-anatomy analysis shows cardiac changes are substantially easier to 
detect than pulmonary changes, while language model comparison reveals 
distinct sensitivity-specificity trade-offs relevant to clinical deployment.

Future work includes incorporating explicit temporal attention mechanisms 
for improved subtle change detection, extending to multi-image sequences 
for longitudinal disease tracking, and developing anatomy-specific models 
to address the challenge of pulmonary change detection.

\bibliographystyle{splncs04}
\bibliography{references}

\end{document}